\documentclass[letterpaper, 10 pt, conference]{ieeeconf}  
\IEEEoverridecommandlockouts

\usepackage{graphicx} 
\usepackage{amsmath} 
\usepackage{amssymb}  

\usepackage{amsthm}
\usepackage{mathtools}
\usepackage{bm}
\usepackage{amsmath}
\usepackage{graphicx}

\usepackage{algorithm}
\usepackage{cite}
\usepackage{algpseudocode}[noend]
\usepackage{multirow}
\usepackage{array}
\usepackage{soul}
\usepackage[dvipsnames]{xcolor}

\usepackage{caption}
\usepackage{subcaption}

\newcommand{\red}[1]{{\normalsize{{\color{red} #1 }}}}

\DeclareMathOperator*{\argmax}{arg\,max}

\usepackage[linkcolor=blue,colorlinks=true]{hyperref}
\usepackage{verbatim}

\theoremstyle{definition}
\newtheorem{definition}{Definition}

\newtheorem{lemma}{Lemma}
\newtheorem{proposition}{Proposition}

\newtheorem{assumption}{Assumption}

\newtheorem{objective}{Objective}
\usepackage{subcaption}
\DeclareCaptionType{equ}[][]

\graphicspath{{cdc2020figures/}} 

\title{\LARGE \bf Algorithms for Finding Compatible Constraints in Receding-Horizon Control of Dynamical Systems 
}

\author{Hardik Parwana$^{1,*}$, Ruiyang Wang$^{1,*}$, and Dimitra Panagou$^{1,2}$ \\ 
\thanks{$^{*}$Both authors have equal contribution.}
\thanks{$^1$ Department of Robotics, $^2$ Department of Aerospace Engineering, University of Michigan, Ann Arbor, MI, USA.}
\thanks{Emails: {\tt\small {hardiksp, ruiyangw, dpanagou}@umich.edu}}
\thanks{This work was partially sponsored by the Office of Naval Research (ONR), under grant number N00014-20-1-2395. The views and conclusions contained herein are those of the authors only and should not be interpreted as representing those of ONR, the U.S. Navy or the U.S. Government.}
}

\newcommand{\reals}{\mathbb{R}}
\newcommand{\s}{\mathcal{S}}
\newcommand{\X}{\mathcal{X}}
\newcommand{\U}{\mathcal{U}}

\newcommand{\K}{\mathcal{K}}

\newcommand{\C}{\mathcal{C}}

\newcommand{\classK}{\mbox{class-$\K$} }

\newcommand{\eqn}[1]{\begin{align}#1\end{align}}

\begin{document}

\maketitle
\thispagestyle{empty}
\pagestyle{empty}


\begin{abstract}

This paper addresses synthesizing receding-horizon controllers for nonlinear, control-affine dynamical systems under multiple incompatible hard and soft constraints. Handling incompatibility of constraints has mostly been addressed in literature by relaxing the soft constraints via slack variables. However, this may lead to trajectories that are far from the optimal solution and may compromise satisfaction of the hard constraints over time. In that regard, permanently dropping incompatible soft constraints may be beneficial for the satisfaction over time of the hard constraints (under the assumption that hard constraints are compatible with each other at initial time). To this end, motivated by approximate methods on the maximal feasible subset (maxFS) selection problem, we propose heuristics that depend on the Lagrange multipliers of the constraints. The main observation for using heuristics based on the Lagrange multipliers instead of slack variables (which is the standard approach in the related literature of finding maxFS) is that when the optimization is feasible, the Lagrange multiplier of a given constraint is non-zero, in contrast to the slack variable which is zero. This observation is particularly useful in the case of a dynamical nonlinear system where its control input is computed recursively as the optimization of a cost functional subject to the system dynamics and constraints, in the sense that the Lagrange multipliers of the constraints over a prediction horizon can indicate the constraints to be dropped so that the resulting constraints are compatible. The method is evaluated empirically in a case study with a robot navigating under multiple time and state constraints, and compared to a greedy method based on the Lagrange multiplier.

\end{abstract}

\section{Introduction}

Autonomous robots are often employed to complete multiple (time-critical and safety-critical) tasks, such as to follow time-stamped waypoints, viewed as soft constraints, while avoiding obstacles, viewed as hard constraints. 
In the case when the tasks are incompatible, one may pose the question of how to relax, or permanently drop them. This problem is in general NP-hard. To address this question, we consider optimization problems with hard and soft constraints, and propose heuristics for removing soft constraints until all the hard constraints are met. 

More specifically, in this paper we consider the control synthesis for nonlinear control-affine systems under a receding-horizon optimization, subject to tasks that are captured by a set of hard and soft time-dependent state constraints. The objective is to maximize a performance score that depends both on safety (satisfaction of hard constraints) and task compatibility (selection of compatible soft constraints). We take a constrained optimization-inspired \cite{marler2004survey,deb2013multi} viewpoint on task selection. When an optimization problem is infeasible, i.e., it does not admit a solution, several methods exist to find the maximal subset of constraints that can be satisfied \cite{chinneck2019maximum}, a problem that is known to be NP-Hard. In general, finding this set has exponential complexity in the number of constraints and therefore heuristics are often proposed \cite{sadegh1999maximum,chinneck2007feasibility}. A related problem is to find a subset of the given cardinality of a set that optimizes a performance metric. This problem is also known to be NP-hard \cite{qian2015subset,davis1997adaptive}.

We propose a framework to decide which soft constraints to be dropped so that the optimization problem for the receding-horizon controller has a feasible solution at every time step. We also show that selecting constraints that help maximize a performance metric for dynamical systems is NP-hard. We first show an application of Chinneck's Algorithm \cite{chinneck1996effective}, a common method based on slack variables in the field of maximal feasible subset selection, to solving our problem. Next, we propose a novel heuristic based on Lagrange multipliers\cite{kirk2004optimal}. When the constraints are deemed incompatible under a given policy, we assign a Lagrange score to each of these constraints based on the variation in the corresponding Lagrange multipliers over the compatible time horizon. These scores are then used to decide the order in which constraints are dropped.
Compared to heuristics with slack variables, our heuristic utilizes information before the optimization problem becomes infeasible. On the contrary, slack variables will only have non-zero values, therefore meaningful information, at the exact time step when the optimization problem is infeasible. To evaluate the efficiency of the proposed heuristic, we compare it to solutions obtained by offline computation of approximate backward reachable sets, and the offline computation of control policies based on optimized Control Barrier Functions (CBFs). 

With regards to the problem considered in our case study, i.e., deciding on tasks (timed waypoints) and generating safe (collision-free and time-respecting) trajectories to these tasks, the literature has considered approaches that are based on optimization and constraint relaxation, however none of the existing methods solves the problem in a systematic manner on how to drop the soft constraints, neither considers nonlinear, control-systems. For example, Model Predictive Control (MPC) and Control Barrier Functions (CBFs) methods can be used to enforce Signal Temporal Logic (STL) specifications for nonlinear system dynamics,  \cite{raman2014model,lindemann2019control,garg2019cdc,garg2022automatica}, however they can not choose or prioritize among constraints in order to prolong the feasibility of the optimization problem. 
With regards to some recent studies that address similar problems where a robot is assigned to travel through a series of regions: \cite{gundana2022event} allows a robot to execute tasks described by Linear Temporal Logic specifications; however, the approach assumes holonomic robots exhibiting straight-line motions, and uses a maximum velocity-based time-to-go estimate as a measure of the reachability of the target region. \cite{notomista2019optimal} and \cite{wang2022time} perform reactive task allocation through relaxed CBF constraints in a CBF-QP controller by weighing the slack variables proportionally to a user-given priority of the constraint. However, all of the aforementioned approaches share the characteristics of a myopic controller, that is, they may enter regions from where the robot may not be able to recover and finish the high-priority tasks. They also do not provide an approach to choose between optional task constraints when they are incompatible with each other or with the other hard constraints imposed by the user. The issue of incompatible safety objectives was recently addressed in \cite{lee2023hierarchical} with a hierarchical relaxation of constraints in the order of absolute priorities defined by the user. However, their approach cannot be extended to cases where priorities have not been defined or are not assigned to individual constraints but to a collection of constraints.

In contrast to the aforementioned studies, our approach, while originally inspired from a time-and-safety-critical task (waypoint) allocation and following problem, offers a much more general framework for compatible constraint selection and hence recursively-feasible control synthesis.
The rest of the paper is structured as follows. Sections  \ref{section::problem_formulation} and \ref{section::methodology} present the problem formulation and proposed methods, respectively, and their evaluation is presented in Section \ref{section::simulation}. 

\section{Preliminaries}

The set of real numbers is denoted as $\reals$ and the non-negative real numbers as $\reals^+$. Given $x\in \reals^n$, $||x||$ denotes $L_2$ norm of $x$. Let $\mathbb{I}(y), y\in \reals$ be step identity function such that $\mathbb{I}(y)=1$ if $y\geq 0$ and $\mathbb{I}(y)=0$ otherwise. The interior and boundary of a set $\C$ are denoted by $\textrm{Int}(\C)$ and $\partial \C$. For $a\in \reals^+$, a continuous function $\alpha:[0,a)\rightarrow[0,\infty)$ is a $\classK$ function if it is strictly increasing and $\alpha(0)=0$. Furthermore, if $a=\infty$ and $\lim_{r\rightarrow \infty} \alpha(r)=\infty$, then it is called class-$\mathcal{K}_\infty$.



Consider a discrete-time nonlinear dynamical system
\eqn{
{x}_{t+1} = f(x_t) + g(x_t)u_t,
\label{eq::dynamics_general}
}
where $x_t\in \X \subset \reals^{n}$ and $u_t \in \U \subset \reals^{m}$ represent the state and control input, and  $f:\X\rightarrow \reals^{n}$ and $g:\X\rightarrow \reals^{m}$ are locally Lipschitz continuous functions. The set of allowable states at time $t$ is specified as an intersection of $N$ sets $\mathcal{S}_i(t),i\in\{1,2,..,N\}$, each of which is defined as the 0-superlevel set of a continuously differentiable function $c_i:\reals^+ \times \mathcal{X} \rightarrow \reals$ as:
    \begin{align}
        \mathcal{S}_{i}(t) & \triangleq \{ x \in \X : c_i(t,x) \geq 0 \}
        \label{eq::safeset}
    \end{align}
    

Define $H_i$: $\reals^+ \times \mathcal{X} \rightarrow \reals$ such that the super-level set of $H_i$ at time $t$, denoted $\mathcal{S}_{H_i}(t) = \{x \in \X : H_i(t,x) \geq 0\}$, defines a constrained set that encodes $n_H$ safety specifications or high-priority tasks that must be completed, where $i\in\mathcal I_H=\{1,\dots,n_H\}$. Similarly, define $h_j$: $\reals^+ \times \mathcal{X} \rightarrow \reals$, where $j\in\mathcal I_h=\{1,\dots,n_S\}$, and denote $\mathcal{S}_{h_j}(t) = \{x \in \X : h_j(t,x) \geq 0\}$ a constrained set that encodes $n_S$ less prioritized tasks that can be removed if needed. Denote $N=n_H+n_S$.

\section{Problem Formulation}
\label{section::problem_formulation}

Let $\pi: \reals^+ \times \mathcal{X} \rightarrow \mathcal{U}$ be a Lipschitz continuous controller that is used by the autonomous agent to enforce safety and performance constraints. Let $\pi^{ref}:\reals^+\times \mathcal{X}\rightarrow \mathcal{U}$ be a reference input controller.

Now given the current state $x_t$ at time $t$, let $x_{\tau|t}$ be the predicted future state at time $\tau+t$ obtained by forward propagation of $x_t$ under dynamics \eqref{eq::dynamics_general} with input $u_{\tau|t}=\pi(\tau+t,x_{\tau|t})$, and reference input $u^{ref}_{\tau|t} = \pi^{ref}(\tau+t, x_{\tau|t})$. Note that $x_{0|t} = x_t$.

This paper specifically considers the controller $\pi$ defined by an optimization problem, termed as the Problem 0 ($\mathrm{P_0}$), of the following form
\begin{subequations}
    \begin{align}
        \textrm{P}_0: \quad & u_{\tau|t} = \pi(\tau+t, x_{\tau|t}) =  \argmax_{u} J(u, \pi_{ref}) \\ 
        \textrm{s.t.} \quad & x_{\tau+1|t} = f( x_{\tau|t} ) + g(x_{\tau|t}) u \label{eq::dynamics_prediction} \\ 
             \nonumber
             & H_i(t+\tau+1, \; x_{\tau+1|t}) \geq 0,               \forall i\in \mathcal{I}_H \\
             & h_j(t+\tau+1, \; x_{\tau+1|t}) \geq 0, \forall j\in \mathcal{I}_h
    \end{align}
    \label{eq::original-problem}
\end{subequations}
where $J(u)$ is a control objective function. The optimization \eqref{eq::original-problem} may not admit a solution for a given time $\tau+t$ and state $x_{\tau|t}$ and in these cases, the controller is said to have failed. 

Suppose that, for a time horizon $T$ of interest, under the flow of dynamics $\eqref{eq::dynamics_general}$, the original optimization problem \eqref{eq::original-problem} becomes infeasible at a prediction time step $t+\tau_{fail}\leq t+T$ and state $x_{\tau_{fail}|t}$. In such cases, to obtain a valid control input, one approach is to find a subset of the soft constraints with indices $\mathcal{I}_h^{f} \subset \mathcal{I}_h$ that give rise to a feasible problem in the whole time horizon when enforced. 

More specifically, consider the following \textbf{Problem 1} $(\mathrm P_1)$, i.e., a reduced problem (called also thereafter a subproblem), where all the hard and a subset of the soft constraints are enforced:
\begin{subequations}
    \begin{align}
        \textrm{P}_1: \quad & u_{\tau|t} =  \quad  \argmax_{u} J(u, u_{ref}) \\ 
        \textrm{s.t.} \quad & x_{\tau+1|t} = f( x_{\tau|t} ) + g(x_{\tau|t}) u \\ 
             \nonumber
             & H_i(t+\tau+1, \; x_{\tau+1|t}) \geq 0,               \forall i\in \mathcal{I}_H \\
             & h_j(t+\tau+1, \; x_{\tau+1|t}) \geq 0, \forall j\in \mathcal{I}_h^{f} (t,x_t),
    \end{align}
    \label{eq::sub-problem}
\end{subequations}
where the set $\mathcal{I}_{h}^{f}(t,x_t)$ at time $t$, at state $x_t$, is such that the optimization problem \eqref{eq::sub-problem} is feasible $\forall \tau \in\{0, 1, 2, \dots, T\}$.

From now on, we use the terms "subproblem" and "reduced problem" interchangeably. 

\begin{definition}
    (Feasible subset) 
    A set $\mathcal{I}_h^f(t,x_t)$ is called a feasible subset at time $t$ and state $x_t$ if the solution $u_{\tau|t}$ of $(\mathrm P_1)$ given out of \eqref{eq::sub-problem} exists $\forall \tau \in \{0, 1, \dots, T\}$.
    \label{definition::feasible_subset}
\end{definition}
A feasible subset $\mathcal{I}_h^{f}(t,x_t)$ can be non-unique, and therefore we search for the feasible subset that furthermore satisfies certain properties. Denote $\mathcal P(\mathcal{I}_h)$ the power set of $\mathcal{I}_h$, and denote $\mathcal F \mathcal P(\mathcal{I}_h)(t,x_t) \subseteq \mathcal P(\mathcal I_h)$ the set that comprises all the elements of $\mathcal P(\mathcal I_h)$ that are feasible subsets of $\mathcal{I}_h$ at time $t$ and state $x_t$. 

Our first objective is to find the feasible subset of maximal cardinality, as defined in Objective \ref{objective::objective1}:
\begin{objective}
    If the set $\mathcal X_{H}(t) = \{x_t \in \mathcal X \;|\; H_i(t,x_t)\geq 0, \forall {i} \in \mathcal{I}_H\}$ is non-empty at time $t$, $\mathcal X_H(t)\neq \emptyset$, find the element of $\mathcal{FP}(\mathcal I_h)(t,x_t)$ of maximal cardinality, denoted $\mathcal I_h^{*f}(t,x_t)$, or in other words, the maximal feasible subset of $(\mathrm P_1)$, defined as:
   \eqn{
    \mathcal I_h^{*f}(t,x_t) = & \argmax \left|\mathcal F \mathcal P(\mathcal{I}_h)(t,x_t)\right|.
   }
   \label{objective::objective1}
More generally, for the case when a reward function of the form  $R:\mathcal P(\mathcal I_h)\rightarrow \mathbb R$ is given, we consider the maximization of the given reward as:
    \eqn{
        \mathcal{I}_h^{*f}(t,x_t)  = & \argmax R(\mathcal F \mathcal P(\mathcal{I}_h)(t,x_t)).}
   \label{objective::objective2}  
   \vspace{-5mm}
\end{objective}
Objective \ref{objective::objective1} is a NP-hard problem. To see why, consider a simple scenario in which $(\mathrm P_0): \pi(\tau+t,x_{\tau|t})$ is feasible at the prediction time $\tau = 0$, but becomes infeasible at time $\tau = 1$. The selection of feasible subsets of soft constraints when considering only $\pi(1+t,x_{1|t})$ is equivalent to the maximum feasible subset problem (maxFS), which is a known NP-hard problem. Objective \ref{objective::objective1} considers not only $\pi(1+t,x_{1|t})$ but also its correlation with $\pi(0+t,x_{0|t})$ when finding feasible subset of soft constraints, therefore it is NP-hard equivalently. 
Specifically, Objective \ref{objective::objective1} is a combinatorial problem with worst-case exponential complexity as the power set $\mathcal P (\mathcal{I}_h)$ has $2^{n_S}$ elements.

\section{Methodology}
\label{section::methodology}
In this section, we propose a heuristic approach to solve for Objective \ref{objective::objective1}. Consider the point-wise optimal QP controller \eqref{eq::original-problem} that enforces $N=n_H+n_S$ (hard and soft) constraints. Let the trajectory at a future time step $\tau$ resulting from forward-propagating the closed-loop system \eqref{eq::dynamics_general} under the controller $ \eqref{eq::original-problem}$ at time $t$, be given by $(x_{\tau|t},u_{\tau|t}, \tau \in \{0,..,\tau_{fail}\})$ where $\tau_{fail}\in \{0,..,\infty\}$ is such that a solution to \eqref{eq::original-problem} exists for all $\tau<\tau_{fail}$ and does not exist for $\tau=\tau_{fail}$. 

\subsection{Chinneck's algorithm}
Several heuristics have been developed for finding the maxFS. In this section, we first briefly review one such iterative algorithm, Chinneck's algorithm, and propose a simple extension of it to our problem. Consider the following optimization problem:
\begin{subequations}
    \eqn{
\min \quad & J(u) \\ 
\textrm{s.t.} \quad &  c_i(u)\leq 0, \;\; i\in \mathcal I_c = \{1,..,N\}.
}
\label{eq::chinneck_opti}
\end{subequations}
and suppose that \eqref{eq::chinneck_opti} is infeasible. The idea is that in each iteration of the algorithm, Chinneck's algorithm removes one constraint permanently until a feasible subset of the constraints is found. More specifically, let the set of constraint indices at iteration $iter\in\{0,1,N-1\}$ be $\mathcal{I}^{iter}$ with $\mathcal I^0 = \mathcal I^c$. At an iterative step with index set $\mathcal{I}^{iter}$, each constraint $k\in\mathcal{I}^{iter}$ is assigned a score, hereby called Chinneck's score, by evaluating the following reduced relaxed problem $\textrm{RRP}(\mathcal I^{iter})_k$ that result by removing $k^{th}$ constraint and adding slack variables to the remaining constraints in \eqref{eq::chinneck_opti}. 
\vspace{-3mm}
\begin{subequations}
    \eqn{
\textrm{RRP}(\mathcal I^{iter})_k: \min \quad & J(u) + M \sum_{i\in \mathcal I^{iter} \setminus \{k\}} \delta_{c_i}^2 \\ 
\textrm{s.t.} \quad &  c_i(u)\leq \delta_{c_i}, \; i\in \mathcal I^{iter} \setminus \{k\}.
}
\label{eq::chinneck_subproblem}
\end{subequations}
where $\delta_{c_i}$ are slack variables, while adding $M\delta_{c_i}^2$ with $M\gg1$ promotes $\delta_{c_i}=0$. The Chinneck's score of constraint $k$ is now taken as the sum of all slack variables $\delta_{c_i}$, $i\in\mathcal I^{iter}\setminus\{k\}$ of $\textrm{RRP}(\mathcal I^{iter})_k$ and is an indication of how much the constraints, $\{c_i \geq \delta_{c_i} \; \forall i \in \mathcal I^{iter} \setminus \{k\}\}$, had to be violated for the $\textrm{RRP}(\mathcal I^{iter})_k$ to admit a feasible solution without relaxation:
\eqn{
r_{chinneck}(\textrm{RRP}(\mathcal I^{iter})_k) = \sum_{i\in \mathcal I^{iter} \setminus \{k\} } \delta_{c_i}. 
}
The constraint $k$ with the largest Chinneck's score $r_{chinneck}(\textrm{RRP}(\mathcal I^{iter})_k)$ is permanently removed at iteration $iter$, and the process is repeated until a feasible solution in which all slack variables are zero is found.




\subsection{Application of Chinneck's Score to Problem P$_0$}

In order to apply Chinneck's method to Problem P$_0$ given by \eqref{eq::original-problem}, we have to account for the fact that we have a dynamical system, i.e., that the control inputs and the states are functions of time (trajectories), and not fixed (constant) with respect to time, and so we perform the following modification: At each iterative step, we first remove one soft constraint $k\in \mathcal{I}^{iter}_h \subset \mathcal{I}_h$ temporarily, which results in a reduced problem (RP) as defined:
\begin{subequations}
    \begin{align}
        \textrm{RP}(\mathcal I^{iter}_h)_k:\quad & u_{\tau|t} =  \quad  \argmax_{u} J(u) \\ 
        \textrm{s.t.} \quad & x_{\tau+1|t} = f( x_{\tau|t} ) + g(x_{\tau|t}) u \\ 
             \nonumber
              H_i(t+\tau+1, \; & x_{\tau+1|t}) \geq 0,               \forall i\in \mathcal{I}_H \\
              h_j(t+\tau+1, \; & x_{\tau+1|t}) \geq 0, \forall j\in \mathcal I^{iter}_h \setminus\{k\}
    \end{align}
    \label{eq::reduced-problem}
\end{subequations}
and propagate the state $x_t$ under dynamics \eqref{eq::dynamics_general} and \eqref{eq::reduced-problem} for $\tau=\{0,1,..,\min(T, \tau_{fail}-1)\}$ where $\tau_{fail}$ is the time at which \eqref{eq::reduced-problem} becomes infeasible. If $\tau_{fail} \leq T$, we evaluate a reduced relaxed problem (RRP) and compute the sum of slack variables of all the hard and the present soft constraints at the failing step $\tau_{fail}$. More specifically, we consider each RRP $(\mathcal I^{iter}_h)_{k}$ given as:
    \begin{align}
        \nonumber
& \textrm{RRP}(\mathcal I^{iter}_h)_k:  u_{\tau_{fail}|t} =  \argmax_{u} J(u) \\ 
        \textrm{s.t.} \quad & x_{\tau_{fail}+1|t} = f( x_{\tau_{fail}|t} ) + g(x_{\tau_{fail}|t}) u \label{eq::sub-problem2}\\ 
             & H_i(t+\tau_{fail}+1, \; x_{\tau_{fail}+1|t}) + \delta_{H_i}\geq 0,               \forall i\in \mathcal{I}_H\nonumber  \\
             & h_j(t+\tau_{fail}+1, \; x_{\tau_{fail}+1|t}) + \delta_{h_j} \geq 0, \forall j\in \mathcal I^{iter}_h \setminus 
             \{k\} \nonumber
    \end{align}
    
and compute a Chinneck score as:
\begin{align}
    & r_{chinneck}(\textrm{RRP}(\mathcal I^{iter})_k))
    = \nonumber\\ 
    & \sum_{\forall i\in \mathcal{I}_H}{\delta_{H_i}} + \sum_{\forall j\in \mathcal I^{iter}_h \setminus k}{\delta_{h_j}} + \frac{1}{R(\mathcal I^{iter}_h \setminus \{k\})}. 
\end{align}
Similarly to the original Chinneck's algorithm, we then permanently drop the constraint whose removal corresponds to a reduced relaxed problem with the largest slack score at each iterative step, and stop until the slack score becomes zero, or all the soft constraints have been removed. A pseudo-code can be found in Algorithm \ref{alg::determine}.

\subsection{Proposed Algorithms based on Lagrange Score}
Chinneck's algorithm utilizes slack variables in the heuristic to find the feasible subsets. However, the slack variables become non-zero only when the optimization problem is infeasible, i.e., at time step $\tau_{fail}$. Therefore, the feasible time steps $\tau\in \{0,..,\tau_{fail}-1\}$ do not contribute to the slack score. However, \eqref{eq::dynamics_general} being a dynamical system, we hypothesize that the time steps preceding failure provide meaningful information about the evolution of the constraint functions over the system trajectories. In accordance with this hypothesis, we propose to use Lagrange multipliers resulting from the optimizations \eqref{eq::sub-problem}, rather than the slack variables.

Our key observation regarding the slack variable and the Lagrange multipliers corresponding to a single constraint is that the former is zero if the problem is feasible, while the latter is not; in that regard, the Lagrange multipliers gives more information about the strictness of the constraint (how much it constrains the feasible space), which is particularly useful when solving for dynamical trajectories over time horizons. It is worth noting that Lagrange multipliers have been known to offer insight into the satisfiability of the constraints and are often used to select candidate constraints that could be made feasible, for example, in the popular active set method \cite{wong2011active} and interior point methods \cite{potra2000interior} for solving quadratic programs.  Our novelty lies in utilizing Lagrange multipliers to evaluate the satisfaction of a constraint over a time horizon for an evolving system such as \eqref{eq::dynamics_general}, and using it to find a feasible subset of constraints. 

Let us first consider the problem \eqref{eq::sub-problem} solved for time horizon $T'=\min(T, \tau_{fail})$ for $\mathcal{I}_h^{'}\subset \mathcal{I}_h$, where $\mathcal I_h'$ is any subset of $\mathcal I_h$. Let $\lambda_{h_j}(\tau)$ be the Lagrange multiplier of the inequality constraint $ h_j(t+\tau+1, x_{\tau+1|t}) \geq 0$ resulting from solving \eqref{eq::sub-problem} at time $\tau$. We assign a \textit{Lagrange value} $l_{h_j}$ to each soft constraint $h_j$ as follows
\eqn{
   l_{h_j} = \sum_{\tau=0}^{\tau_{fail}-1} \lambda_{h_j}(\tau). 
   \label{eq::largrange_score}
}
We conjecture that the impact a constraint has on the feasibility of \eqref{eq::sub-problem} over a time horizon is proportional to its Lagrange value, and therefore we remove the constraint with the highest $l_{h_j}$. In what follows, we will describe two heuristics utilizing the Lagrange values of the constraints to compute feasible subsets.

\subsubsection{Proposed Method on Finding Feasible Subproblems based on Lagrange Multipliers}

Similar to Chinneck's algorithm, we present an iterative procedure to remove constraints successively. Consider the relaxed subproblem \eqref{eq::reduced-problem} that results from removing constraint $k \in \mathcal{I}^{iter}_h$ at iteration $iter$. We forward propagate the state $x_{\tau}$ under \eqref{eq::dynamics_general} and \eqref{eq::reduced-problem} for a time horizon $T'=\min(T, \tau_{fail})$. The Lagrange score of constraint $k$ is designed as follows
\begin{align}
    & r_{lag}(\mathrm{RP}(\mathcal I^{iter}_h)_k) = \nonumber \\
    &\begin{cases}
      \sum^{\mathcal I^{iter}_h \setminus\{k\}}_j l_j  + \frac{1}{R(\mathcal I^{iter}_h \setminus\{k\})} \quad \textrm{if } \tau_{fail} \leq T \\
      \frac{1}{R(\mathcal I^{iter}_h \setminus\{k\})} \qquad \qquad \qquad \;\; \text{otherwise}
    \end{cases}
\label{eq::r_lag_values}
\end{align}
Now we remove the constraint with largest Lagrange score and the procedure continues until we find a feasible solution or remove all soft constraints as described in Algorithm \ref{alg::determine}.

\subsubsection{A Greedy Approach based on Lagrange Multipliers}
The approach in Algorithm \ref{alg::determine} requires solving optimization problems for multiple subproblems in each iteration. This procedure can have a high computational overhead that might not be suitable for real-time implementation. In this section, we present a greedy approach
Algorithm \ref{alg:greedy} to select a subset of soft constraints by dropping them in the order of Lagrange value. Specifically, consider the subproblem \eqref{eq::sub-problem} solved for a constraint set $\mathcal{I}_h^{iter}\subset \mathcal{I}_h$ and $\tau_{fail}$ be the time \eqref{eq::sub-problem} first becomes infeasible. Then we assign the following score to each constraint $k$
\eqn{
  r_{lag} (\mathrm{P}_1(\mathcal I^{iter}_h)_k) = l_{h_k}
}
and permanently drop the constraint with the largest score. We repeat this procedure until a feasible solution is found ($\tau_{fail}>T$) or all the soft constraints are dropped.

\subsubsection{Complexity Analysis}

For the subproblem-based algorithm, the complexity of the number of optimizations to be solved in the worst-case, corresponding to the case when all soft constraints are removed, is $O(n_S^2 T)$ because in every iteration $iter \in \{0,..,n_S-1\}$, $\tau \in \{0,..,T-1\}$, it needs to solve $n_S-1$ optimizations. The overall computational complexity also depends on the solve time for each optimization \eqref{eq::original-problem}. For quadratic programs, for example, polynomial time solvers are available\cite{stellato2020osqp} (although QP, being NP-Hard in general, have exponential complexity). For such solvers, the complexity of the subproblem approach is $O(n_S^4 T)$.
As for the greedy algorithm, since it drops one constraint at each step and checks whether the reduced problem is feasible or not, its worst-case complexity of the number of solved optimizations is $O(n_ST)$, and computational complexity for quadratic programs is $O(n_S^3 T )$.

\begin{algorithm}
\begin{algorithmic}[1]
\caption{Feasible Subproblem Algorithm}
\State  $\mathcal I^0_h = \{1,..,n_S\} = \mathcal I_h$ \Comment{Constraint Indices to consider}
\State $\mathcal I^f_h = \emptyset$ 
\Comment{Feasible Index Set Initialization}
\For{$iter \in \{0,1,\dots,n_S-1\}$}
    \State $r_{min} = INF$
    \If {Chinneck}
        \State $r_k = r_{chinneck} (\mathrm{RRP}(\mathcal I^{iter}_h)_k)$
    \Else
        \State $r_k =  r_{lag}(\mathrm{RP}(\mathcal I^{iter}_h)_k)$
    \EndIf
     \If{$r_k < r_{min}$}
            \State $r_{min} = r_k$, $k_{drop} = k$
    \EndIf
    \State $\mathcal I^{iter+1}_h = \mathcal I^{iter}_h \setminus k_{drop}$
    \If{$r_{min} == 0$}
        \State $\mathcal I^f = \mathcal I^{iter+1}$, break
    \EndIf
\EndFor
\State return $\mathcal I^f_h$
\label{alg::determine}
\end{algorithmic}
\end{algorithm}

\begin{algorithm}
\begin{algorithmic}[1]
\caption{Greedy Algorithm}
\State $x_0$ \Comment{Current state of the system }
\State $\mathcal I^0_h = \{1,..,n_S\} = \mathcal I_h$ \Comment{Constraint Indices to consider}
\State $\mathcal I^f_h = \emptyset$ \Comment{Feasible Index Set Initialization}

\For{$iter \in \{0, n_S-1\}$}
    \State $k_{max} = \argmax_{k}(r_{lag} (\mathrm{P}_1(\mathcal I^{iter}_h)_k))$
    \If{$\mathrm P_1(\mathcal I^{iter}_h)$ is feasible}
    \State $\mathcal I^f_h = \mathcal I^{iter}_h$, break
    \EndIf
    \State $\mathcal I^{iter+1}_h = \mathcal I^{iter}_h \setminus k_{max}$
\EndFor
\State return $\mathcal I^f_h$
\label{alg:greedy}
\end{algorithmic}
\end{algorithm}

\section{Simulation Results}
\label{section::simulation}
\begin{figure}
    \centering\includegraphics[width=0.75\linewidth]{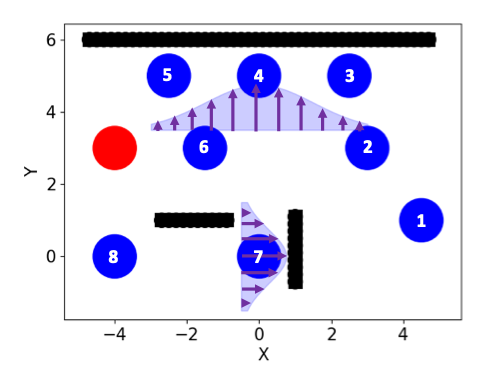}
    \caption{\small{A sample testing environment, in which each waypoint is shown in blue and the shaded blue regions show disturbances affecting the robot. The robot is required to travel through the series of waypoints in a specified order (as shown in white numbers) and returns to the red final position before $T_{final}$}}
     \label{fig:motivation}   
     \vspace{-5mm}
\end{figure}

\begin{figure*}[h!]

\subfloat[Case 1]
{   
    \includegraphics[width=0.33\linewidth,height=4.5cm]{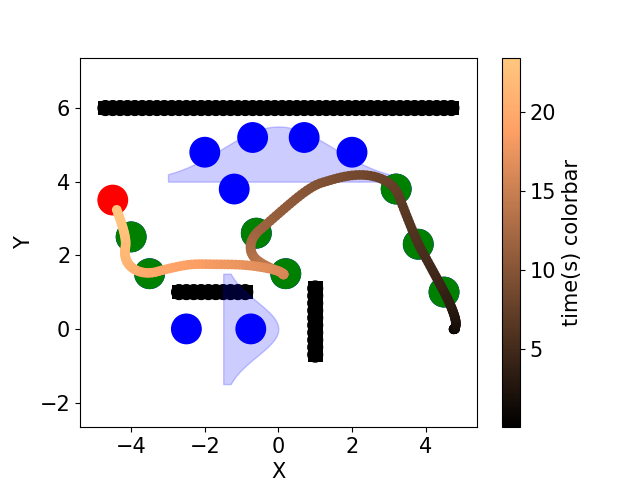}
}
\subfloat[Case 2]
{  
    \includegraphics[width=0.33\linewidth,height=4.5cm]{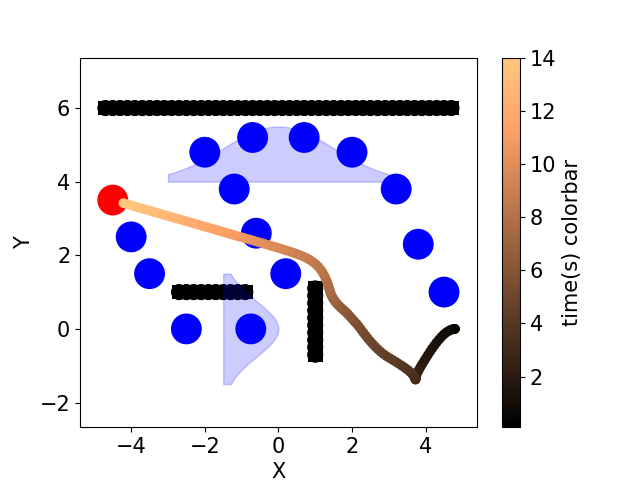}
}
\subfloat[Case 3]
{   
    \includegraphics[width=0.33\linewidth,height=4.5cm]{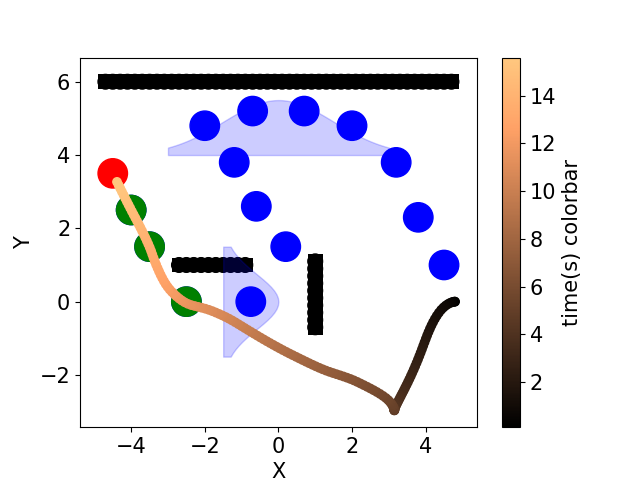}
}
\caption{\small{Testing environment with 14 waypoints. Each waypoint is rewarded and their reward is shown in Table \ref{table::sim2}. The trajectory calculated from (a) Case 1: Exhaustive CBF Search (Optimal), (b) Case 2: Algorithm 2 with $r_{chinneck}$, (c) Case 3: Algorithm 2with $r_{lag}$ under the reward Case 1. 
}}
\label{fig:test2}
\end{figure*}

In this section, we illustrate the efficacy of our algorithm through simulations. Consider the scenario shown in Figs.~ \ref{fig:motivation}, \ref{fig:test1} and \ref{fig:test2}, in which a robot is tasked to track a series of time-stamped waypoints, i.e., the robot is required to arrive at the $j$-th waypoint before a user-defined time $T_j$. The strict requirements on the robot are avoiding collision with the walls and reaching the final target position within user-defined time $T_{final}$. Let the robot position at time $t$ be denoted $p_t$. The constraints to reach a waypoint $p_{j}$ and the final target position $p_{tf}$ at maximum allowable distance of $d_{max}$ are formulated through the following Control Lyapunov functions (CLFs) \cite{sontag1989universal}:
\eqn{
 V_{j}(t,p_t) = d_{max}^2 - (p_{j} - p_{t})^T(p_{j}- p_{t}) \\
    V_{tf}(t,p_t) = d_{max}^2 - (p_{tf} - p_t)^T(p_t-p_{t}).
}
The constraints for the waypoints are encoded via the CLF conditions:
\begin{align}
     \nonumber
     h_j(t,p_t)
     = \phi_{j}(t)\Bigg[V_{j}(t+1,p_{t+1})- (1+\alpha_{h_j}) V_{j}(t,p_t)\Bigg]
\end{align}
$\forall j\in \mathcal I_h(t)=\{1,\dots,n_{waypoints}\}$, where $\alpha_{h_j}>0$ and 
$\phi_{j}(t)$ is the activation function defined as:
\eqn{
   \phi_{j}(t) = 
        \begin{cases}
            1, \text{if }  T_{j-1}<t<T_{j}\\
            0, & \text{otherwise}
        \end{cases}
}
where $T_j$ is a user-defined time specification so that the robot has to arrive at the $j$-th waypoint before $T_j$. When we remove waypoints, we shift the lower bound of the activation function of each included waypoint to the upper bound of the activation function of the included waypoint just before it accordingly. As a result, $[T_{j-1}, T_{j}]$ is the active time window of the $j$-th waypoint.

Similarly, the hard constraint for arrival at the final target is defined as:
\begin{align}
     \nonumber
     H_{tf}(t,p_t) = \phi_{tf}(t)\Bigg[V_{tf}(t+1,p_{t+1})- (1+\alpha_{H_{tf}}) V_{tf}(t,p_t)\Bigg]
\end{align}
where $\alpha_{H_{tf}}>0$, and $\phi_{tf}$ is defined with a final time $T_{final}$ so that the robot is required to arrive at the final target position before $T_{final}$.
Collision avoidance with static obstacles, which are approximated as a series of circular disks, is formulated using the following barrier function 
\eqn{
b_{Hi}(t,p_t) = (p_t - x_{oi})^T(p_t- x_{oi}) - d^2_{min}
}
and the Control Barrier Function (CBF)\cite{ames2016control, lindemann2018control} condition
\eqn{
    H_i(t,p_t) = b_{Hi}(t+1,p_{t+1}) - (1+\alpha_{Hi}) b_{Hi}(t,p_t)
}
$\forall i \in \mathcal I_{H}(t) = \{1,\dots,n_{obstacles}\}$, 
where $x_{oi}$ is the position of the centroid of the $i$-th circular static obstacle, and $d_{min}$ is the radius of the obstacle.

The CLF-CBF-QP\cite{ames2016control} controller that is employed is
\begin{align}
        \min_{u_{\tau|t}\in \reals^m} \quad & J(u_{\tau|t}) = (u_{\tau|t}-u_{ref,\tau|t})^T(u_{\tau|t}-u_{ref,\tau|t}) \nonumber   \\
        \textrm{s.t.} \quad & h_j (\tau|t,x_{\tau+1|t}) \geq 0, ~j\in \{1,..,n_{waypoints}\}\nonumber  \\
        & H_i (\tau|t,x_{\tau+1|t}) \geq 0, ~ i \in \{  1,..,n_{obstacles}  \} \label{eq::CBF_Single}\nonumber\\
        & H_{tf} (\tau|t,x_{\tau+1|t}) \geq 0, \; u_{\tau|t} \leq 1.0  
\end{align}
 
where $u_{ref,\tau|t}$ is the reference control signal at time $\tau+t$, designed to steer the robot towards the active waypoint.

In our implementation, the QP controller \eqref{eq::CBF_Single} is used to generate control inputs for the robot, which is modeled either under single-integrator dynamics (Test 1) or under the dynamic unicycle (Test 2) dynamics.


The environment is affected by known disturbances, whose magnitude is represented with light blue color in Figs. \ref{fig:motivation}, \ref{fig:test1}, \ref{fig:test2}. The nonlinear disturbance may lead to a collision in the presence of input bounds if the robot navigates to one of the shaded areas. Thus, the robot may have to drop some of the soft constraints corresponding to blue waypoints to be able to reach the last red waypoint while remaining safe. In both test cases, the original problem, 
P$_0$, defined in (\ref{eq::CBF_Single}), is rendered infeasible within the initial time interval 
$[0,T]$, where $T$ is the controller’s prediction time horizon. This infeasibility arises when mandating arrival at every waypoint with the presence of disturbances, necessitating the removal of some waypoints at 
$t=0$. We evaluated all results on Intel® Core™ i7-9700 CPU @ 3.00 GHz with Python 3.8.10. 
\subsubsection{Test 1}
\begin{figure*}
\hfill
\subfloat[Failure with slack variables]
{   
    \includegraphics[width=0.33\linewidth,height=4.5cm]{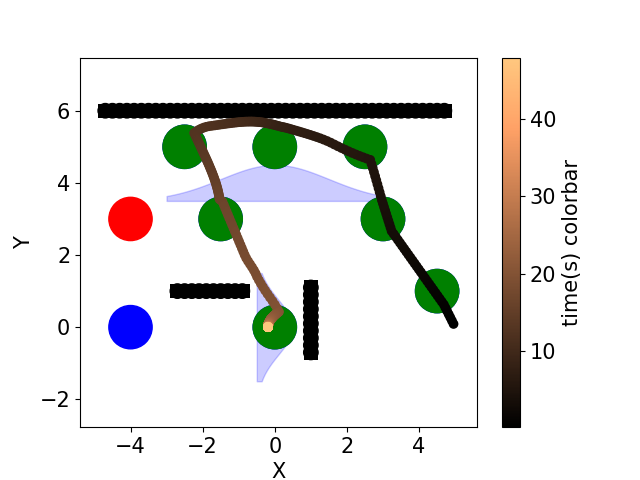}
}
\subfloat[Case 1]
{   
    \includegraphics[width=0.33\linewidth,height=4.5cm]{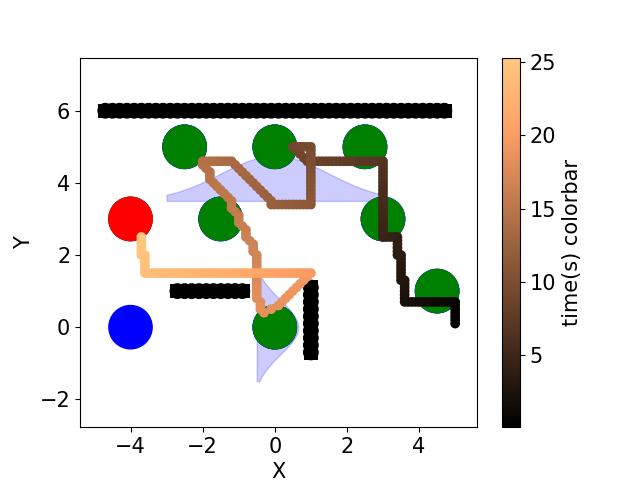}
}
\subfloat[Case 2]
{  
    \includegraphics[width=0.33\linewidth,height=4.5cm]{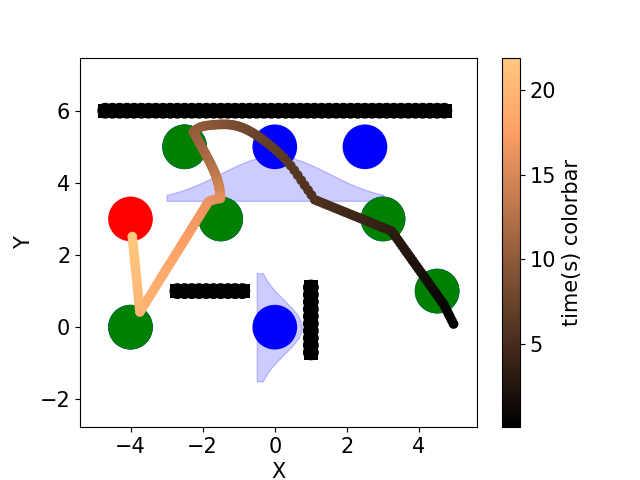}
}

\hfill
\subfloat[Case 3]
{   
    \includegraphics[width=0.33\linewidth,height=4.5cm]{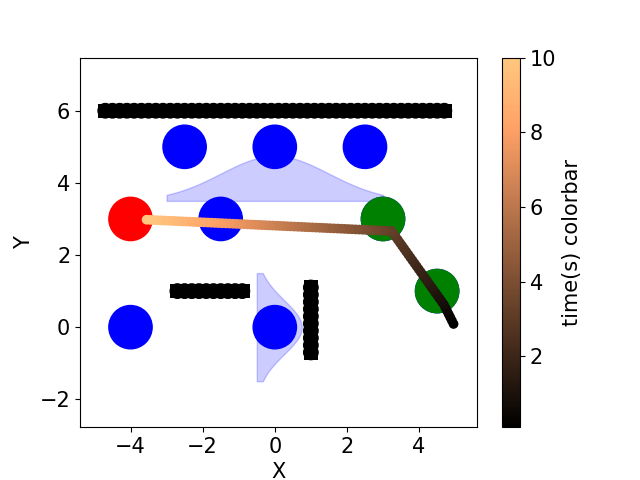}
}
\subfloat[Case 4]
{   
    \includegraphics[width=0.33\linewidth,height=4.5cm]{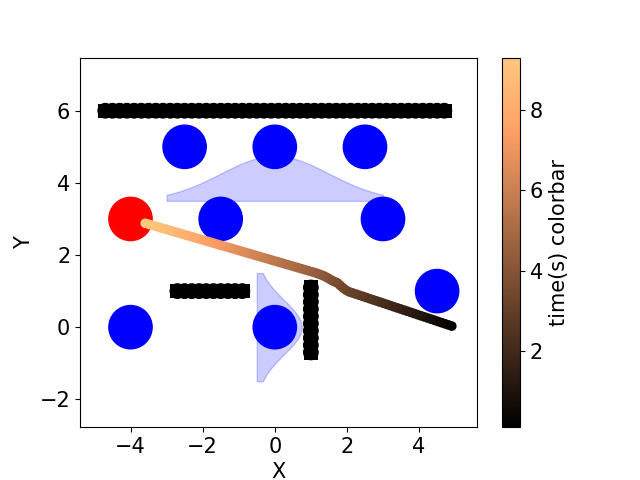}
}
\subfloat[Case 5]
{      \includegraphics[width=0.33\linewidth,height=4.5cm]{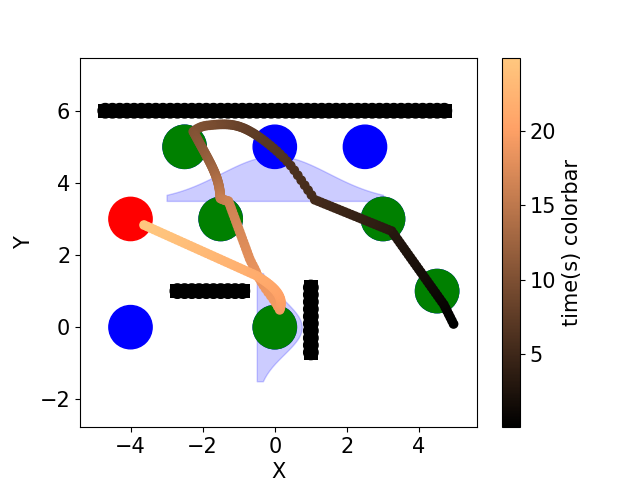}}

\caption{\small{(a) Relaxing soft constraints: The robot converges to a deadlock (caused by the superposition of disturbances and constraints), and fails to reach the red waypoint within the given time. The trajectory calculated from (b) Case 1: Reachability, (c) Case 2: Exhaustive CBF Search, (d) Case 3: Greedy Algorithm with  $r_{lag}$, (d) Case 4: Subproblem Algorithm with $r_{chinneck}$, (e) Case 5: Subproblem Algorithm with $r_{lag}$. Note that the waypoints 4 in (c), 6 in (d), and 4 in (f), were dropped by the controller as they could not be reached within the prescribed time. The paths just happen to pass through them while the robot aims the next waypoint.}}
\label{fig:test1}
\end{figure*}

We first start with each waypoint being rewarded equally, and we compare the number of waypoints arrived between different algorithms. The prediction time horizon in this scenario is equal to $T_{final}=250$ (equivalent to 25s in real-time). This paper primarily seeks to underscore the advantages of selectively removing soft constraints over the common method of adding slack variables when facing infeasibility issues. Fig. \ref{fig:test1}(a) illustrates that while introducing slack variables allows the robot to reach waypoints, disturbances can prevent it from reaching the final target within the user-defined time, thus violating the hard constraints. In contrast to this case, our proposed method of removing selected soft constraints, as will be illustrated through Figs. \ref{fig:test1}(b)-(f) and \ref{fig:test2}, ensures that the robot meets all safety-related hard constraints and successfully reaches the target.

Next, we present results with 7 different testing environments under three different numbers of waypoints and under three different disturbance levels in TABLE \ref{table::sim1}. The "Offline" columns of the Table include the methods we implemented for comparison, namely: A "Reachability" approach, where the optimal solution is obtained by computing backward reachable sets as mentioned in Appendix \ref{section::reachability}, and an "Exhaustive CBF search" approach, where the reward $R(\mathcal P(\mathcal{S}_h)_f)$ is designed to maximize the number of waypoints reached, and the Objective \ref{objective::objective1} is solved using an exhaustive search by evaluating all possible QP controllers of the form \eqref{eq::CBF_Single} with different combinations of soft constraints at every time. The results of our methods are shown in the "Heuristic" columns. Namely, we show the results of the "Greedy" heuristic that utilizes the Lagrange score $r_{lag}$, of Chinneck's method that utilizes Chinneck's heuristic score $r_{chinneck}$, and of the subproblem method that utilizes the Lagrange score $r_{lag}$, respectively. We apply all the heuristics at the initial time and find a subset of constraints that can be rendered feasible for all time in the future (specifically till $T_{final}$). Once such a subset is found, the robot implements the receding-horizon controller with this subset of constraints and the trajectories with 8 waypoints and the Medium disturbance level from all cases are shown in Fig. \ref{fig:test1}. 

\begin{table*}[h!]
\centering
\begin{tabular}{|m{4em}|m{3em}|m{4em}|m{3em}|m{4em}|m{3em}|m{4em}|m{3em}|m{4em}|m{4em}|m{4em}|m{3em}|}
      \hline
      \multicolumn{2}{|m{7em}|}{} &\multicolumn{4}{m{14em}|}{OPTIMAL} & \multicolumn{6}{m{21em}|}{HEURISTICS} \\
      \hline
      No. Waypoints & Disturb Level & \multicolumn{2}{m{7em}|}{Reachability} & \multicolumn{2}{m{7em}|}{Exhaustive CBF Search} & \multicolumn{2}{m{7em}|}{Algorithm 2 based on $r_{lag}$} & \multicolumn{2}{m{8em}|}{Algorithm 1 based on $r_{chinneck}$} & \multicolumn{2}{m{7em}|}{Algorithm 1 based on $r_{lag}$} \\
      \hline
      \multicolumn{2}{|m{7em}|}{} &No. Waypoints Reached & Comp. Time & No. Waypoints Reached & Comp. Time & No. Waypoints Reached & Comp. Time & No. Waypoints Reached & Comp. Time  & No. Waypoints Reached & Comp. Time\\
      \hline
      \multirow{3}{*}{6} & Low and Medium & \multirow{3}{*}{6} & \multirow{3}{*}{5.55h} & 6 & 1.74s &\multicolumn{6}{m{29em}|}{Every CBF methods reaches 6 waypoints with the same computational time because the original problem is feasible with CBF based controller} \\
      \cline{2-2}
      \cline{5-12}
      & High & & & 4 & 20.86s & 1 & 2.58s & 2 & 8.87s & 2 & 8.56s \\
      \hline
      \multirow{3}{*}{8} & Low & 8  & \multirow{3}{*}{5.55h} & 7 & 560.6s & 7 & 4.64s & 7 & 23.24s & 7 & 22.69s \\
      \cline{2-3}
      \cline{5-12}
      & Medium & 7 & & 5 & 236.3s & 2 & 4.14s & 0 & 19.64s & 5 & 15.04s  \\
      \cline{2-3}
      \cline{5-12}
      & High & 5 & & 4 & 156.2s & 2 & 3.63s & 1 & 17.47s & 2 & 19.79s \\
      \hline
      \multirow{3}{*}{14} & Low & 14 & \multirow{3}{*}{5.55h} & 12 & 3.61h & 12 & 6.95s & 5 & 234.32s & 9 & 164.01s  \\
      \cline{2-3}
      \cline{5-12}   & Medium & 13 & & 9 & 2.61h & 1 & 7.01s & 2 & 55.52s & 3 & 48.54s \\
      \cline{2-3}
      \cline{5-12}      & High & 12 & & 8 & 2.55h & 2 & 6.34s & 0 & 52.18s & 2 & 47.19s  \\
      \hline
\end{tabular}
\caption{\small{Number of waypoints reached with different algorithms under the single integrator dynamics.}}
\label{table::sim1}
\vspace{-2mm}
\end{table*}

The computation time is also summarized in TABLE \ref{table::sim1}. The calculation time for "Reachability" is of the order 20000 seconds. The number of subproblems required in the exhaustive search for the best CBF controller grows exponentially in the number of constraints, thus its calculation increases dramatically with an increased number of waypoints. On the other hand, the subproblem algorithm with both heuristics and the greedy algorithm only needs to solve a linear number of subproblems with respect to the number of all soft constraints.

Based on the results in TABLE \ref{table::sim1}, we observe that the subproblem algorithm with either $r_{chinneck}$ or $r_{lag}$ heuristic can arrive at more waypoints compared to the greedy algorithm, but at the cost of solving more subproblems, thus taking a longer computational time. We also observe the expected improved performance in terms of more waypoints arrived when replacing the heuristic from $r_{chinneck}$ to $r_{lag}$.
Furthermore, the subproblems that the subproblem algorithm needed to solve grow linearly with the number of soft constraints, which is more efficient than the exhaustive search method, which needs to solve an exponentially increasing number of subproblems with an increase in number of soft constraints.


\subsubsection{Test 2}
For our second testing scenario, we considered an environment with 14 waypoints and "High" disturbance level, as defined in Test 1.
The time horizon for our controller in this scenario is also equal to $T_{final}$ = 250. 

Shown in TABLE \ref{table::sim2} are results from simulations with 3 different sets of rewards for each waypoint. We evaluate the proposed algorithm with both the $r_{lag}$ and $r_{chinneck}$ heuristic in solving \textbf{Objective \ref{objective::objective2}}. 

\begin{table}[h!]
\centering
\begin{tabular}{|m{2.7em}|m{2.5em}|m{2.5em}|m{3em}|m{4em}|m{3em}|m{3em}|}
      \hline
       & \multicolumn{2}{m{5em}|}{OPTIMAL} & \multicolumn{4}{m{12em}|}{HEURISTICS} \\
      \hline
      Reward & \multicolumn{2}{m{5em}|}{Exhaustive CBF Search} & \multicolumn{2}{m{7.5em}|}{Algorithm 1 based on $r_{chinneck}$ } & \multicolumn{2}{m{6em}|}{Algorithm 1 based on $r_{lag}$}\\
      \hline
       &Reward & Time & Reward & Time & Reward & Time \\
      \hline
      Case 1 & 17 & 7.22h & 0 & 123.08s & 12 & 118.49s\\
      \hline
      Case 2 & 18 & 6.32h & 0 & 124.31s & 3 & 119.02s \\
      \hline
      Case 3 & 12 & 6.33h & 0 & 123.56s & 5 & 119.25s \\
      \hline
\end{tabular}
\caption{\small{Reward achieved and computation time for different algorithms. rewards for each waypoint are as follows: Case 1: [1,1,1,1,1,2,2,2,2,2,4,4,4,4], Case 2: [4,4,4,4,4,2,2,2,2,2,1,1,1,1], Case 3: [1,2,3,4,1,2,3,4,1,2,3,2,1,2].
}}
\label{table::sim2}
\end{table}



Based on results shown in TABLE \ref{table::sim2}, our proposed heuristic with the $r_{lan}$ score performs better compared to Chinneck's algorithm with $r_{chinneck}$ score in terms of being able to reach more waypoints than Chinneck's method, at a lower computational effort. Both methods are also much faster than an exhaustive search over all possible combinations of constraints. 

We observe that Lagrange variables are well suited to reason about the feasibility of optimization problems for a dynamical system. We conjecture that a possible reason for the good performance of our heuristic is that the optimization is more likely to become infeasible if it has more active constraints in preceding time steps. Since Lagrange multipliers are non-zero only for active constraints, they are a measure of 'total activity'. Further theoretical analysis is left for future work.

\section{Conclusion}
We developed heuristics for finding feasible sets of constraints by dropping soft constraints based on the  Lagrange multipliers of each constraint over a time horizon. Evaluation of our algorithm is then done for a test case of robot waypoint following in the presence of obstacles and disturbances. Future work will involve the formal analysis of the convergence guarantees of the proposed heuristic. Finally, to improve the optimality of our solutions, we also plan to induce randomness into our decision-making by using approaches such as genetic algorithms whose fitness score could be made a function of our heuristic.

\bibliographystyle{IEEEtran}
\bibliography{cdc.bib}

\appendix
\subsection{Dynamics Definition}
\subsubsection{Single Integrator Dynamics}
\label{sec::single_dynamics} Let state and control inputs: $x = \begin{bmatrix} p_x & p_y\end{bmatrix}^T$, $u = \begin{bmatrix}v_x & v_y\end{bmatrix}^T$, the dynamics is: $x_{t+1} = x_{t} + u_t\; dt$, 
with $dt$ being the discrete time step.

\subsubsection{Dynamic Unicycle Dynamics}
\label{sec::unicycle_dynamics}
For the Dynamic Unicycle dynamics used, we define state and control inputs: $x = \begin{bmatrix}x & y & v & \theta\end{bmatrix}^T$, $u = \begin{bmatrix}a & \omega\end{bmatrix}^T$, and their dynamics as $x_{t+1} = x_{t} + \begin{bmatrix}v_t \cos(\theta_t) & v_t \sin(\theta_t) & a_t & \omega_t\end{bmatrix}^T dt$, where $dt$ is the discrete time step.
 
\subsection{Reachability based Solution}
\label{section::reachability}
In the waypoint tracking scenarios discussed in section \ref{section::simulation}, a reachability-based solution is employed. 
Let the backward reachable set $\mathcal{BRS}(p_{tf})$ of final target waypoint $p_{tf}$ under dynamics \eqref{eq::dynamics_prediction} is given by
\begin{align}
    \nonumber
    \mathcal{BR}_{p_{tf}}(t) &= \{x ~|~ x_{0|t} = x, \;
    \exists u_{\tau|t},\\ \textrm{ s.t. } & H_{i}(t+\tau, x_{\tau|t}) \geq  0, \forall \tau\in \{0,1,.., T\}\}
\end{align}
Also define the unsafe backward reachable set $\mathcal{UBR}_i(t)$  under dynamics \eqref{eq::dynamics_prediction} of each unsafe region as:
\begin{align}
    \nonumber
    \mathcal{UBR}_i(t) &= \{x ~| ~x_{0|t} = x, \forall u_{\tau|t}, \; \exists \tau_{unsafe}, \\ 
    \nonumber
    &  \textrm{ s.t. }  H_{i}(x_{\tau_{unsafe}|t},\tau_{unsafe}|t) < 0\}
\end{align}
This computation is amenable to offline computation only, even for simple systems \cite{bansal2017hamilton}. 
At time $t$ and state $x_t$, the robot chooses to go the next waypoint $j'$ if there exists a path $p(t)$ going to the waypoint $j'$ with $p(0)=x_t$ such that 
\eqn{
  p(t) \in \mathcal{BRS}(p_{tf}) \land p(t) \notin \mathcal{UBR}_{Hi}(t) ~ \forall t>0.
}
To implement the above decision-making process, we first discretize the state and control input space into a grid and make a graph connecting grid cells such that each edge represents a feasible transition between cells in one-time step. Subsequently, Dijkstra's algorithm is used for each waypoint combination to ascertain the existence of a safe path, enabling the timely reaching of each waypoint and the target position, with path length verifying the time requirements for reaching each waypoint.

\end{document}